\pdfoutput=1
\documentclass[11pt]{article}
\usepackage{textgreek}
\usepackage{algorithm}
\usepackage[noend]{algpseudocode} 
\usepackage{amsmath}
\usepackage[x11names]{xcolor}
\usepackage{amssymb}
\usepackage{enumitem}

\usepackage[final]{acl}

\usepackage{times}
\usepackage{latexsym}

\usepackage[T1]{fontenc}

\usepackage[utf8]{inputenc}

\usepackage{microtype}

\usepackage{inconsolata}

\usepackage{graphicx}
\usepackage{booktabs}
\usepackage{multirow}
\usepackage[table]{xcolor}
\definecolor{mygr}{RGB}{0,128,0}
\usepackage{xparse}

\usepackage{tabularx}
\usepackage{amsthm}
\usepackage{soul}

\makeatletter
\newsavebox{\UPB@box}
\newenvironment{UnifiedPromptBox}[1]{%
  \par\noindent\begingroup
  \def\UPB@title{#1}%
  \setlength{\fboxrule}{0.8pt}%
  \setlength{\fboxsep}{0pt}%
  \begin{lrbox}{\UPB@box}%
  \begin{minipage}{\linewidth}%
    {\setlength{\fboxsep}{3mm}%
     \colorbox{black!80}{%
       \parbox{\dimexpr\linewidth-2\fboxsep\relax}{\color{white}\bfseries \UPB@title}%
     }%
    }\par\vspace{2mm}%
    \small\setlength{\parindent}{0pt}%
    \leftskip=3mm \rightskip=3mm
    \ignorespaces
}{%
    \par%
  \end{minipage}%
  \end{lrbox}%
  \fcolorbox{black!80}{white}{\usebox{\UPB@box}}%
  \endgroup\par\medskip
}
\makeatother

\title{HAG: Hierarchical Demographic Tree-based Agent Generation \\ for Topic-Adaptive Simulation}

\author{
    Rongxin Chen$^{1,2}$, 
    Tianyu Wu$^{1,2}$, 
    Bingbing Xu$^{1}$\thanks{\; Corresponding author.}, \\ 
    \textbf{Jiatang Luo}$^{1,3}$, 
    \textbf{Xiucheng Xu}$^{1,2}$, 
    \textbf{Huawei Shen}$^{1,2}$\\
        $^{1}$State Key Laboratory of AI Safety, Institute of Computing Technology, CAS, \\
        $^{2}$University of Chinese Academy of Sciences \\
        $^{3}$School of Advanced Interdisciplinary Sciences, University of Chinese Academy of Sciences \\
        \texttt{\{chenrongxin24s,wutianyu25s,xubingbing\}@ict.ac.cn}
}

\begin{document}
\maketitle
\begin{abstract}
High-fidelity agent initialization is crucial for credible Agent-Based Modeling across diverse domains. A robust framework should be Topic-Adaptive, capturing macro-level joint distributions while ensuring micro-level individual rationality.
Existing approaches fall into two categories: static data-based retrieval methods that fail to adapt to unseen topics absent from the data, and LLM-based generation methods that lack macro-level distribution awareness, resulting in inconsistencies between micro-level persona attributes and reality.
To address these problems, we propose HAG, a Hierarchical Agent Generation framework that formalizes population generation as a two-stage decision process. 
Firstly, utilizing a World Knowledge Model to infer hierarchical conditional probabilities to construct the Topic-Adaptive Tree, achieving macro-level distribution alignment. Then, grounded real-world data, instantiation and agentic augmentation are carried out to ensure micro-level consistency.
Given the lack of specialized evaluation, we establish a multi-domain benchmark and a comprehensive PACE evaluation framework. Extensive experiments show that HAG significantly outperforms representative baselines, reducing population alignment errors by an average of 37.7\% and enhancing sociological consistency by 18.8\%.  Code and dataset are released at 
\url{https://github.com/Libra117/HAG}.

\end{abstract}

\section{Introduction}

\begin{figure}[t]
  \centering
  \includegraphics[width=1\linewidth]{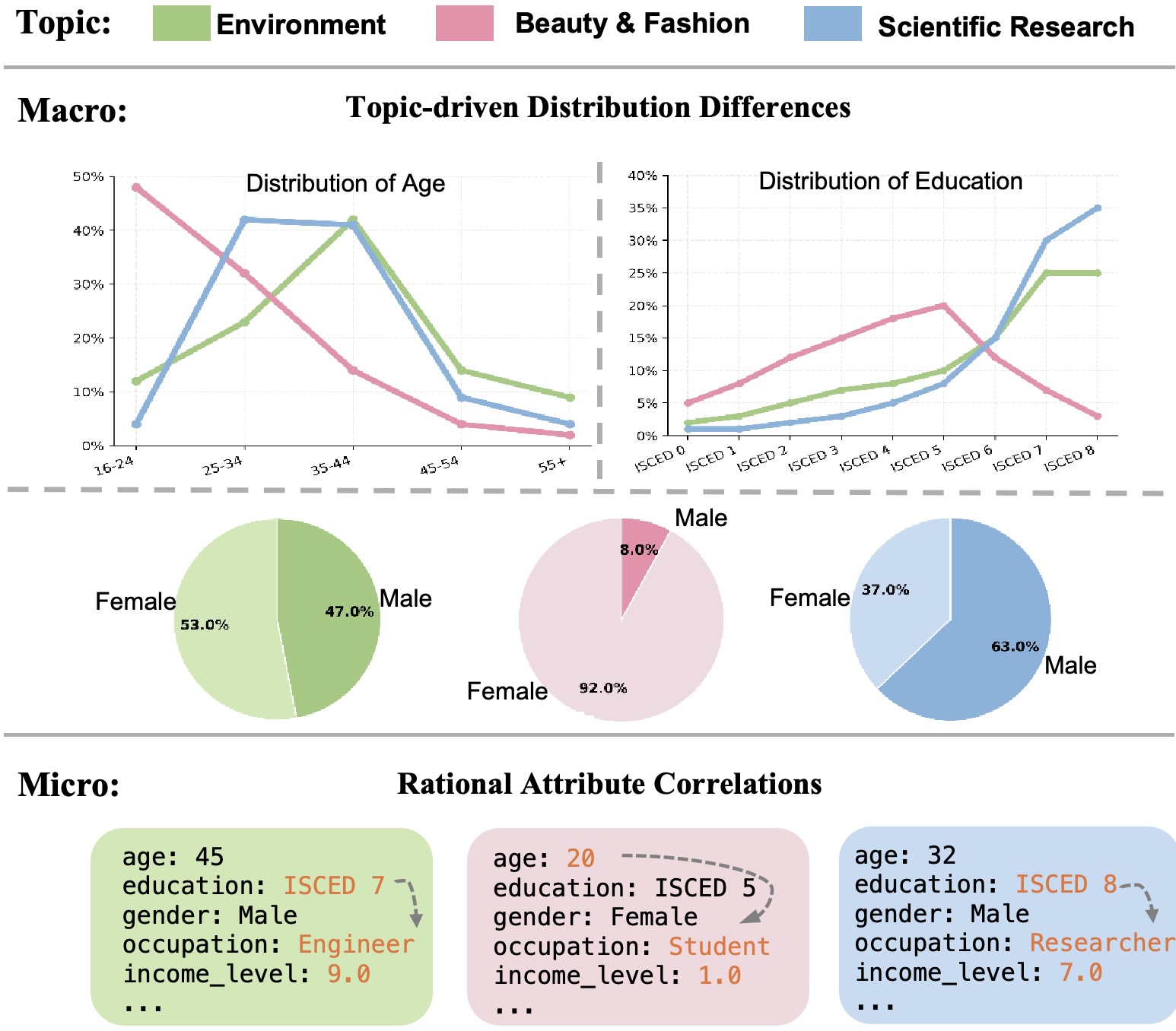}
  \caption{The necessity of Topic-Adaptive in simulation.}
  \label{fig.1}
\end{figure}

With the advancement of Large Language Models (LLMs), agent-based modeling (ABM) plays a key role in simulating complex multi-agent interactions, with growing adoption in computational social science~\cite{abm_social, manning2024automated, wangsurvey}, economic modeling~\cite{abm_econ, ecoagent}, and personalized recommendation~\cite{abm_recsys, recsyssurvey}.
These simulations share a common requirement: heavily depend on User Agents to simulate preferences and interactions. Therefore, the fidelity of such a simulation system is essentially limited by the quality of agents, whether the simulated population accurately represents the demographic heterogeneity and behavioral differences of real world~\cite{fidelity, population_simu}.

A robust agent generation framework for simulation should be fundamentally Topic-Adaptive. As illustrated in Figure~\ref{fig.1}, demographic structures and attribute distributions are not static but changing across different topics. Capturing these topic-specific differences requires satisfying two critical criteria: at the macro level, the generated population should model the correct joint distribution of multi-dimensional attributes conditioned on the topic, ensuring the overall demographic persona aligns with the specific scenario rather than treating agents as isolated individuals; and at the micro level, individual agents should present rational and coherent attribute correlations, reflecting realistic sociological dependencies rather than random combinations. Failure to meet these criteria inevitably leads to significant divergence in simulation trajectories and unreliable downstream analysis.

Existing agent generation approaches~\cite{autoagentsurvey, agent_syn, persona_survey}, however, struggle to fulfill these criteria. They fall into two categories: data-based retrieval and LLM-based generation. Data-based retrieval constructs agent pools directly from real-world user logs~\cite{socioverse, oasis, usp, datadriven}, but are inherently static and tightly coupled to historical data, limiting their ability to adapt to unseen or data-scarce topics. LLM-based generation methods construct agent personas via predefined schemas~\cite{yulan, promise_catch, llmgenerate} or text-based inference~\cite{persona_hub, pop_aligned}. While more flexible, they are typically constrained by sparse expert knowledge and construct population by aggregating independent individuals, lacking explicit modeling of the joint distributions over multi-dimensional attributes and multi-agents~\cite{distribution, socialagenteval}. 
As a result, when lacking real data support, individuals may exhibit incongruous persona attributes, rendering simulations ineffective~\cite {incongruous_persona, validation, promise_catch, noise}.
Overall, to date, no existing approach simultaneously achieves both topic-adaptive population macro-level modeling and micro-level rationality.

To combat the challenges, we propose \textbf{HAG}, a \textbf{H}ierarchical \textbf{A}gent \textbf{G}eneration framework that formalizes population generation as a hierarchical decision process. By encoding the target topic as a latent guiding prompt, HAG achieves progressive persona generation in two stages:
1) Topic-Adaptive Tree Construction: Utilizing a World Knowledge Model, the framework autonomously derives node values and path weights by computing hierarchical conditional probabilities across salient demographic dimensions. This replaces manual heuristics with automated expert priors, ensuring macro-distributions are contextually aligned with the target topic.
2) Grounded Instantiation \& Agentic Augmentation: To populate the tree, the framework grounds leaf nodes via real-world persona retrieval. For data-deficient nodes, we employ agentic augmentation to synthesize missing personas, satisfying global macro-constraints while maintaining micro-level consistency.
In summary, HAG enables the efficient generation of diverse, context-aware, and sociologically grounded agent populations for high-fidelity simulations.

Given the lack of specialized evaluation for agent generation, we establish a multi-domain benchmark, spanning social simulation, product recommendation, and movie critique, and we propose \textbf{PACE} (\textbf{P}opulation \textbf{A}lignment \& \textbf{C}onsistency \textbf{E}valuation). This framework provides comprehensive metrics to quantitatively assess the generated population from two complementary dimensions: statistical alignment and sociological consistency.
Extensive experiments show that HAG significantly outperforms representative baselines, reducing population alignment errors by an average of \textbf{37.7\%} and enhancing sociological consistency by \textbf{18.8\%}. Code and benchmark are released.

\begin{figure*}[ht]
\centering
  \includegraphics[width=1\linewidth]{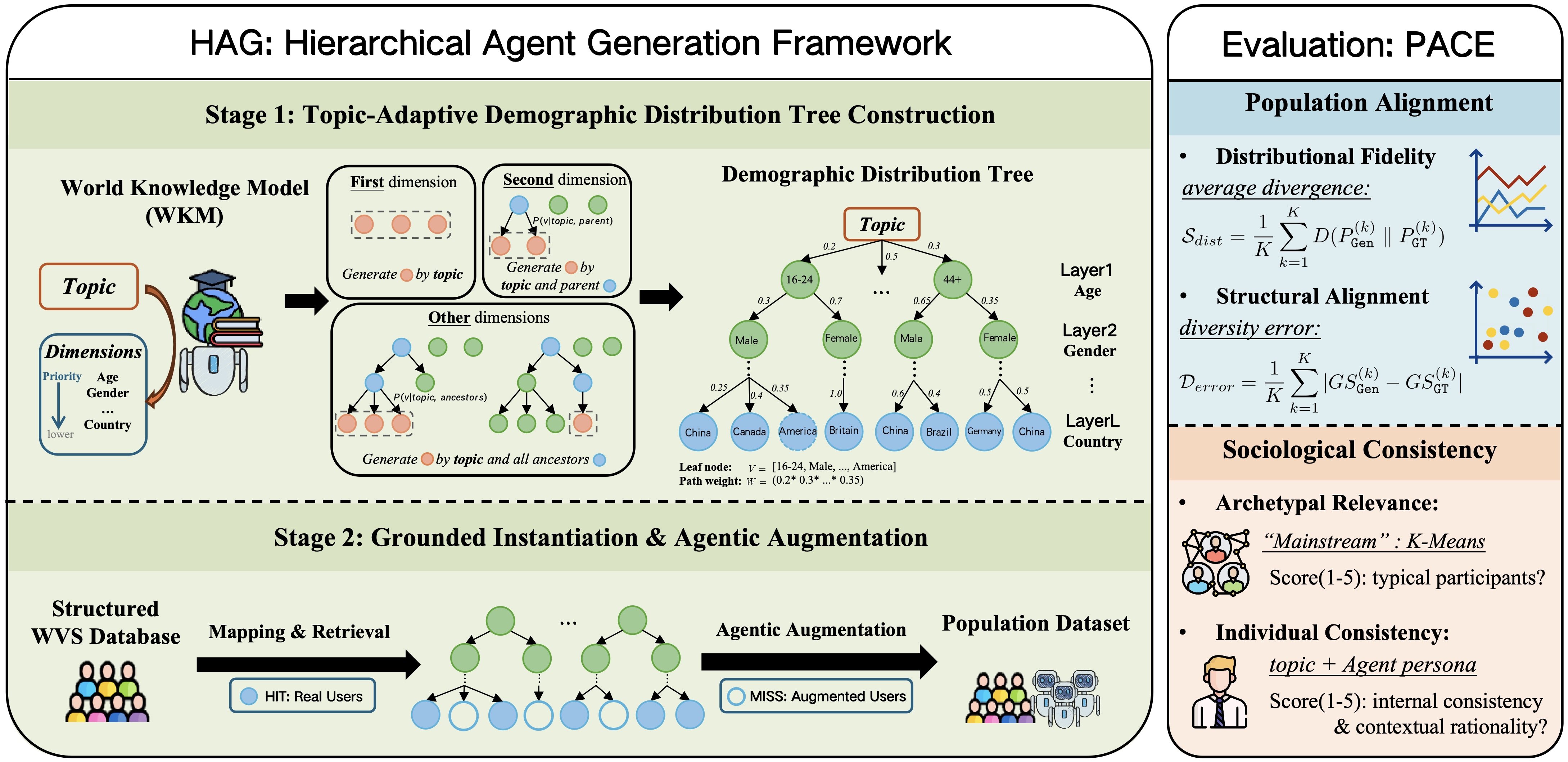} 
  \caption {Illustration of the HAG framework. It utilizes the World Knowledge Model to construct a Topic-Adaptive Demographic Distribution Tree, and generates population based on the tree via filtering real users and agentic data augmentation. Evaluation of the generated population is from two aspects: population alignment and sociological consistency.}
  \label{fig:method}
\end{figure*}

\section{Methodology}

In this section, we propose \textbf{HAG}, a \textbf{H}ierarchical Demographic Tree-based \textbf{A}gent \textbf{G}eneration framework for Topic-Adaptive Simulation. 
We first provide a formal definition of the topic-adaptive population generation problem, and then detail the HAG framework. An illustration of our framework is shown in the left part of Figure~\ref{fig:method}.

\subsection{Problem Definition}

We consider the task of constructing a topic-based agentic population dataset grounded in real demographic data. For a given input topic $t$ and a target population size $N$, our goal is to generate an agent population $\mathcal{P}=\{p_j\}_{j=1}^{N}$ whose demographic composition matches the target demographic structure of that topic. Each agent $p_j$ is associated with a persona vector $\mathbf{x}_j = \{v^{(1)}, v^{(2)}, \dots, v^{(L)}\}$ defined over a set of $L$ selected demographic dimensions (e.g., Age, Gender), and each $v^{(i)}$ represents the specific value of the agent on the $i$-th dimension.

The core challenge lies in the fact that the ideal demographic composition for a specific topic is often latent. We denote this target persona distribution as $\mathbf{D}(t)$, which represents the distribution of demographic attributes given the topic $t$. 
At the macro level, the real distribution of the generation dataset, denoted as $\widehat{\mathbf{D}}_{\mathcal{P}}$, should approximate the theoretical target $\mathbf{D}(t)$, capturing the correct conditional dependencies between attributes rather than treating them independently. At the micro level, each generated individual $\mathbf{x}_j$ should fall within the valid real-world data, avoiding unrealistic attribute combinations. Consequently, our objective is to derive a structured representation of $\mathbf{D}(t)$ and instantiate a concrete population dataset that satisfies both topic-adaptivity and grounding constraints.

\subsection{HAG Framework}

The HAG framework achieves this objective through two stages: (1) constructing a hierarchical demographic distribution tree to capture topic-relevant demographic dimensions and their joint probabilities of values. (2) grounding this tree in real-world data by filtering real users and augmenting generated agents where data is missing. This effectively bridges the gap between macro-level distribution modeling and micro-level individual consistency.

\subsubsection{Topic-Adaptive Demographic Distribution Tree Construction}

To explicitly model the macro-level joint distribution, we represent the topic-based population structure as a hierarchical demographic distribution tree $\mathcal{T}_d$.

For a given topic $t$, we use a World Knowledge Model (WKM) to identify and order a subset of relevant demographic dimensions from a pre-specified set $\mathcal{F}$. Formally, the WKM outputs a prioritized dimension sequence:
\begin{equation}
\begin{aligned}
\mathbf{f}_{\mathrm{prior}}(t)
&=\operatorname{PrioritizeDims}(\mathcal{M},\,t,\,\mathcal{F})\\
&=\bigl(f^{(1)},\ldots,f^{(L)}\bigr)
\end{aligned}
\end{equation}
where $\mathcal{M}$ denotes the WKM, $\operatorname{PrioritizeDims}(\cdot)$ is the function to identify and order topic-relevant dimensions. This sequence directly determines the layer order of the distribution tree from root to leaves, where $f^{(1)}$ represents the first dimension with the highest priority and $f^{(L)}$ has the lowest priority among the selected dimensions.

The tree is constructed top-down along this ordered sequence driven by the WKM. With the input topic $t$ as the root node, the demographic dimensions are encoded in hierarchical order of priority, with higher priority dimensions expanding earlier.
At the first layer, nodes are generated directly conditioned on the topic $t$. For any subsequent layer $l$, nodes are generated based on both the topic and the path of ancestor nodes. Specifically, each node represents a distinct value $v^{(l)}$ of dimension $f^{(l)}$, and each edge encodes a conditional weight. For a parent node representing a partial persona $\bigl(v^{(1)}, \ldots, v^{(l-1)}\bigr)$, the normalized weight of the edge to a child value $v^{(l)}$ is defined as:
\begin{equation}
\begin{aligned}
&w\bigl(v^{(l)} \mid v^{(1:l-1)}, t\bigr)\\
=& P\Bigl(f^{(l)} = v^{(l)} \mid f^{(1:l-1)} = v^{(1:l-1)}, t\Bigr)
\end{aligned}
\end{equation}
where $\sum_{v^{(l)} \in \mathcal{V}(f^{(l)})}w\bigl(v^{(l)} \mid v^{(1:l-1)}, t\bigr) = 1$. This formulation ensures that the probability of each attribute is not static but dynamically conditioned on the preceding context, capturing the dependencies between dimensions.

Finally, each leaf node corresponds to a complete demographic persona $\mathbf{v} = (v^{(1)}, \ldots, v^{(L)})$. The target proportion for this persona is obtained by the product of edge weights along the root-to-leaf path:
\begin{equation}
W(\mathbf{v}\mid t) = \prod_{l=1}^{L} w\bigl(v^{(l)} \mid v^{(1:l-1)}, t\bigr)
\end{equation}
Each $W(\mathbf{v}\mid t)$ shows the proportion of the population with persona $\mathbf{v}$ under topic $t$.
The resulting tree $\mathcal{T}_d$ provides an explicit, structured macro-constraint for population instantiation, translating the abstract topic into a concrete target joint distribution.

\subsubsection{Grounded Instantiation and Agentic Augmentation}
To ensure the generated agents match the distribution of the real-world society rather than hallucinated statistics, we utilize the \textbf{World Values Survey (WVS)}\footnote{\url{https://www.worldvaluessurvey.org/WVSDocumentationWV7.jsp}} as a source of our database. Following the demographic framework established in \textit{WorldValuesBench} \cite{worldvaluesbench}, we extract a comprehensive set of attributes to measure a user's sociological positioning.
Concretely, we filter the raw survey data to select 12 key dimensions classified into three categories: Basic Demographics, Socio-Economic Status, and Cultural Identity. Shown in Table \ref{tab:wvs_attributes}, this classification captures the social attributes of a persona, enabling us to model complex social distributions rooted in real users.

Based on the demographic distribution tree, we construct the final dataset $\mathcal{P}$ by grounding each leaf distribution in this real-world data. Given a total target size $N$, we compute the required count for each leaf persona as $n(\mathbf{v}) = \operatorname{Round}\bigl(N \cdot W(\mathbf{v}\mid t)\bigr)$, where $\operatorname{Round}(\cdot)$ is the rounding function, with minor adjustments to preserve the sum $N$. We then perform a retrieval-and-augmentation process:

For each leaf persona $\mathbf{v}$, we filter the processed WVS database to retrieve matching real users. Let $m(\mathbf{v})$ denote the count of available real users. We assign a coverage status $\mathrm{Tag}(\mathbf{v})$:
\begin{equation}
\mathrm{Tag}(\mathbf{v})=
\begin{cases}
\texttt{HIT}, & m(\mathbf{v}) \ge n(\mathbf{v}),\\
\texttt{MISS}, & m(\mathbf{v}) < n(\mathbf{v}).
\end{cases}
\end{equation}
For \texttt{HIT} nodes, we sample $n(\mathbf{v})$ real users directly, ensuring maximum realism. For \texttt{MISS} nodes, where data scarcity occurs, we sample all available $m(\mathbf{v})$ and generate the remaining deficiency $n(\mathbf{v}) - m(\mathbf{v})$ via agentic augmentation. Crucially, this augmentation is constrained by the specific path $\mathbf{v}$ defined by the tree, preventing the LLM from hallucinating incompatible attribute combinations.

Finally, we obtain the final agentic population dataset $\mathcal{P}$ consisting of filtered real-world users and augmented agents, matching the distribution tree. The detailed procedure is presented in Algorithm~\ref{alg:HAG}.
In summary, HAG effectively solves the mode collapse and "Frankenstein" agent problems by deriving the overall structure from the topic-adaptive tree and preferentially retrieving real data.

\begin{algorithm}[t]
\caption{HAG algorithm}
\label{alg:HAG}

\definecolor{ForestGreen}{RGB}{34,139,34}
\newcommand{\AlgStep}[1]{\Statex {\color{ForestGreen}\textbf{\textit{#1}}}}

\begin{algorithmic}[1]
\Require Topic $t$, Wold Knowledge Model $\mathcal{M}$, Database $\mathcal{D}$, Size $N$, Dimensions $\mathcal{F}$
\Ensure Agent Population $\mathcal{P}$

\AlgStep{Stage 1: Topic-Adaptive Tree Construction}
\State $\mathbf{f} \gets \textsc{PrioritizeDims}(\mathcal{M}, t, \mathcal{F})$
\State Initialize tree $\mathcal{T}$ with root $r$

\Function{Expand}{$u, l, \mathbf{v}_{path}$}
    \If{$l > |\mathbf{f}|$} \Return \EndIf
    \State $\{(v, w)\} \gets \textsc{Infer}(\mathcal{M}, t, \mathbf{f}[l] \mid \mathbf{v}_{path})$
    \For{$(v_i, w_i) \in \{(v, w)\}$}
        \State $c \gets \textsc{Node}(v_i)$
        \State \textsc{AddEdge}($u \to c, w_i$)
        \State \Call{Expand}{$c, l{+}1, \mathbf{v}_{path} \cup \{v_i\}$}
    \EndFor
\EndFunction
\State \Call{Expand}{$r, 1, \emptyset$}

\AlgStep{Stage 2: Grounded Instantiation}
\State $\mathcal{P} \gets \emptyset$
\For{each leaf persona $\mathbf{v}$ in $\mathcal{T}$}
    \State $n_{tgt} \gets \textsc{Round}(N \cdot \textsc{PathProb}(\mathbf{v}))$
    \State $\mathcal{P}_{hit} \gets \textsc{Retrieve}(\mathcal{D}, \mathbf{v}, \text{limit}=n_{tgt})$
    \State $n_{gap} \gets \max(0, n_{tgt} - |\mathcal{P}_{hit}|)$
    \State $\mathcal{P}_{miss} \gets \textsc{Augment}(\mathcal{M}, \mathbf{v}, n_{gap})$
    \State $\mathcal{P} \gets \mathcal{P} \cup \mathcal{P}_{hit} \cup \mathcal{P}_{miss}$
\EndFor
\State \Return $\mathcal{P}$
\end{algorithmic}
\end{algorithm}

\section{Benchmark and Evaluation Framework}
This section constructs a multi-domain benchmark and proposes PACE, a framework designed to quantify generation quality from statistical alignment to sociological consistency.

\subsection{Benchmark Construction}

A critical challenge in evaluating population simulation is the absence of the benchmark that explicitly maps specific topics to fine-grained user demographic distributions. Existing corpus typically contain user-generated content, but lack the structured demographic labels necessary for ground-truth comparison. To address this problem, we construct a Topic-based Benchmark by inferring reference populations directly from behavioral data.

We select three publicly available diverse corpus to cover heterogeneous domains: \textbf{Bluesky Social Dataset}~\footnote{\url{https://zenodo.org/records/14669616}} for social simulation, \textbf{Amazon Reviews 2023}~\footnote{\url{https://huggingface.co/datasets/McAuley-Lab/Amazon-Reviews-2023}} for product recommendation, and \textbf{IMDB User Reviews}~\footnote{\url{https://www.kaggle.com/datasets/sadmadlad/imdb-user-reviews}} for movie critique. We identify representative topics (e.g., discussion themes, product categories, or movies) and aggregate relevant user posts.

To derive the ground-truth demographics for each topic, we employ a \textbf{text-to-persona} pipeline~\cite{persona_hub}. Based on the sociological premise that language patterns reflect latent identity, this pipeline infers demographic personas from user texts integrated from social media posts and comments on things using the ability of LLM. 
This forms a population benchmark under different topics, serving as a reference standard for evaluating the generated population. We also conducted effective manual sampling verification on the constructed population benchmark, showing a consistency rate of 92\%. Detailed information on data subsampling, filtering criteria, and topic selection is provided in Appendices~\ref{sec:appendix_eval_details}.

\subsection{PACE Evaluation Framework}HAG
To measure the quality of generated population, we propose \textbf{PACE }(\textbf{P}opulation \textbf{A}lignment \& \textbf{C}onsistency \textbf{E}valuation), which is structured around two aspects: Population Alignment, which quantifies statistical fidelity against ground truth (GT), and Sociological Consistency, which evaluates the sociological rationality of semantics. An overview of the PACE framework is shown in the right part of Figure~\ref{fig:method}.

\subsubsection{Population Alignment}
This aspect measures the objective statistical distance between the generated population ($P_{\mathtt{Gen}}$) and the real-world ground truth ($P_{\mathtt{GT}}$).

\paragraph{Distributional Fidelity.}
We evaluate the gap between the demographic composition of the generated population and real world with a statistical method. For each attribute dimension $k$, we use two divergences: the Jensen-Shannon divergence (JSD) and the Kullback-Leibler divergence (KL). The overall fidelity score is defined as the average divergence across all $K$ dimensions:
\begin{equation}
    \mathcal{S}_{dist} = \frac{1}{K} \sum_{k=1}^{K} D(P_{\mathtt{Gen}}^{(k)} \parallel P_{\mathtt{GT}}^{(k)})
\end{equation}
where $P_{\mathtt{Gen}}^{(k)}$ and $P_{\mathtt{GT}}^{(k)}$ denote the marginal probability distributions of the $k$-th attribute in the generated and ground truth population, respectively. $D(\cdot \parallel \cdot)$ denotes the divergence metric (JSD or KL), and a lower $\mathcal{S}_{dist}$ indicates higher fidelity to match fundamental demographic proportions.

\paragraph{Structural Alignment.}
In addition to considering marginal distributions, we measure the degree of topic diversity error. We use the Gini-Simpson Index ($GS = 1 - \sum p_i^2$) to quantify diversity, where $p_i$ denotes the proportion of a category $i$~\cite{gini_simpson}, and define the Diversity Error as the absolute difference between the structural diversity of the generated cohort and the GT:
\begin{equation}
    \mathcal{D}_{error} = \frac{1}{K} \sum_{k=1}^{K} | GS_{\mathtt{Gen}}^{(k)} - GS_{\mathtt{GT}}^{(k)} |
\end{equation}
A $\mathcal{D}_{error}$ close to zero indicates that the framework adaptively aligns with the topic's social structure, which preserves diverse topics and narrows the focus for niche discussions.

\subsubsection{Sociological Consistency}
This aspect extends statistics to the sociological semantic quality of the agents.

\paragraph{Archetypal Relevance.}
We evaluate the sociological relevance of the generated “mainstream” voices to the topic. We apply $K$-Means clustering on the agent embeddings to extract the top-$K$ centroids that keep the dominant archetypes. 
Then we evaluate the centroids to determine if they represent typical participants in the specific topic.

\paragraph{Individual Consistency.}
We scrutinize the logical soundness of individual agents through a dual-lens assessment. We conduct a comprehensive evaluation across the entire generated population, ensuring that every generated persona is checked for both internal self-consistency that avoids attribute contradictions, and contextual rationality that ensures plausibility within the specific topic.

\begin{table*}[t]
\centering
\caption{Results of experiments across Bluesky, Amazon, and IMDB domains. Metrics for Population Alignment (JSD, KL, DivErr) are \textit{lower-is-better}, while metrics for Sociological Consistency (ArchRel, IndCon) are \textit{higher-is-better}. "Average " denotes the mean score of the four models.}
\label{tab:main_results}

\providecommand{\uline}[1]{\underline{#1}}
\providecommand{\sout}[1]{%
  \begingroup
  \setbox0=\hbox{#1}%
  \rlap{\raisebox{0.55ex}{\rule{\wd0}{0.4pt}}}#1%
  \endgroup
}
\newcommand{\LightBlueRow}{\rowcolor[RGB]{221,235,247}}
\newcommand{\GrayRow}{\rowcolor[RGB]{242,242,242}}

\setlength{\tabcolsep}{4pt}
\renewcommand{\arraystretch}{1.05}
\resizebox{\textwidth}{!}{%
\begin{tabular}{llccccccccccccccc}
\toprule
\multirow{3}{*}{\textbf{Model}} & \multirow{3}{*}{\textbf{Method}} &
\multicolumn{5}{c}{\textbf{Bluesky Social Dataset}} &
\multicolumn{5}{c}{\textbf{Amazon Reviews 2023}} &
\multicolumn{5}{c}{\textbf{IMDB Movies User Reviews}} \\
\cmidrule(lr){3-7}\cmidrule(lr){8-12}\cmidrule(lr){13-17}
& &
\multicolumn{3}{c}{Alignment $\downarrow$} & \multicolumn{2}{c}{Consistency $\uparrow$} &
\multicolumn{3}{c}{Alignment $\downarrow$} & \multicolumn{2}{c}{Consistency $\uparrow$} &
\multicolumn{3}{c}{Alignment $\downarrow$} & \multicolumn{2}{c}{Consistency $\uparrow$} \\
\cmidrule(lr){3-5}\cmidrule(lr){6-7}
\cmidrule(lr){8-10}\cmidrule(lr){11-12}
\cmidrule(lr){13-15}\cmidrule(lr){16-17}
& &
JSD & KL & DivErr & ArchRel & IndCon &
JSD & KL & DivErr & ArchRel & IndCon &
JSD & KL & DivErr & ArchRel & IndCon \\
\midrule
\multicolumn{17}{l}{\textbf{Data-based Retrieval Methods}} \\
\LightBlueRow \sout{\phantom{N/A}} & Random Select & 0.628 & 2.489 & 0.505 & 3.000 & 2.599 & 0.530 & 1.286 & 0.518 & 3.000 & 2.878 & 0.510 & 1.359 & 0.535 & 2.500 & 3.440 \\
\LightBlueRow \sout{\phantom{N/A}} & Topic-Retrieval & 0.578 & 5.725 & 0.285 & 3.250 & 2.928 & 0.587 & 4.035 & \textbf{0.408} & 3.000 & 3.134 & 0.576 & 2.049 & 0.473 & \uline{3.000} & 3.242 \\
\multicolumn{17}{l}{\textbf{LLM-based Generation Methods}} \\
\LightBlueRow Average & LLM Generate & 0.539 & 2.487 & 0.466 & 3.063 & 3.197 & 0.451 & 0.925 & 0.504 & 2.750 & \textbf{3.675} & 0.479 & \textbf{1.041} & 0.555 & 2.875 & \uline{3.690} \\
\LightBlueRow \strut & HAG-Flat & \uline{0.401} & \uline{2.436} & \uline{0.276} & \uline{3.750} & \uline{3.324} & \uline{0.429} & \uline{0.779} & 0.439 & \uline{3.125} & 3.487 & \uline{0.398} & 3.392 & \uline{0.324} & \uline{3.000} & 3.686 \\
\LightBlueRow \strut & HAG(Our) & \textbf{0.345} & \textbf{1.657} & \textbf{0.263} & \textbf{3.813} & \textbf{3.617} & \textbf{0.414} & \textbf{0.759} & \uline{0.419} & \textbf{3.250} & \uline{3.642} & \textbf{0.393} & \uline{1.331} & \textbf{0.322} & \textbf{3.125} & \textbf{3.783} \\

\GrayRow GPT-4 & LLM Generate & 0.559 & 1.432 & 0.541 & 3.250 & 2.947 & 0.515 & 1.185 & 0.566 & 2.000 & 3.228 & 0.502 & 1.176 & 0.575 & 3.000 & 3.565 \\
\GrayRow \strut & HAG-Flat & 0.401 & 1.497 & 0.295 & 3.750 & 3.315 & 0.421 & 0.784 & 0.432 & 3.000 & 3.547 & 0.379 & 0.914 & 0.417 & 3.000 & 3.785 \\
\GrayRow \strut & HAG(Our) & 0.354 & 1.084 & 0.281 & 3.750 & 3.510 & 0.447 & 0.815 & 0.423 & 3.000 & 3.628 & 0.385 & 0.734 & 0.342 & 3.000 & 3.828 \\

\GrayRow GPT-oss-120b & LLM Generate & 0.485 & 0.967 & 0.468 & 3.250 & 3.363 & 0.409 & 0.688 & 0.473 & 3.000 & 3.685 & 0.461 & 0.884 & 0.521 & 3.000 & 3.619 \\
\GrayRow \strut & HAG-Flat & 0.411 & 2.518 & 0.226 & 3.250 & 3.089 & 0.454 & 0.895 & 0.439 & 3.000 & 3.543 & 0.304 & 0.473 & 0.326 & 3.000 & 3.999 \\
\GrayRow \strut & HAG(Our) & 0.320 & 0.619 & 0.250 & 4.000 & 3.712 & 0.407 & 0.754 & 0.434 & 3.000 & 3.643 & 0.297 & 0.421 & 0.272 & 3.500 & 4.048 \\

\GrayRow Gemini-2.5-Pro & LLM Generate & 0.627 & 6.513 & 0.400 & 2.500 & 2.911 & 0.417 & 0.765 & 0.442 & 3.000 & 4.022 & 0.481 & 1.010 & 0.556 & 3.000 & 3.890 \\
\GrayRow \strut & HAG-Flat & 0.451 & 4.665 & 0.320 & 3.750 & 3.317 & 0.423 & 0.743 & 0.412 & 3.500 & 3.399 & 0.434 & 7.231 & 0.226 & 3.000 & 3.507 \\
\GrayRow \strut & HAG(Our) & 0.393 & 4.052 & 0.259 & 3.500 & 3.558 & 0.401 & 0.738 & 0.431 & 3.000 & 3.619 & 0.423 & 1.034 & 0.350 & 3.000 & 3.652 \\

\GrayRow DeepSeek-V3.2 & LLM Generate & 0.485 & 1.035 & 0.456 & 3.250 & 3.565 & 0.465 & 1.061 & 0.536 & 3.000 & 3.765 & 0.474 & 1.093 & 0.568 & 2.500 & 3.685 \\
\GrayRow \strut & HAG-Flat & 0.343 & 1.063 & 0.264 & 4.250 & 3.576 & 0.418 & 0.695 & 0.474 & 3.000 & 3.457 & 0.476 & 4.950 & 0.326 & 3.000 & 3.454 \\
\GrayRow \strut & HAG(Our) & 0.313 & 0.873 & 0.260 & 4.000 & 3.690 & 0.402 & 0.727 & 0.388 & 4.000 & 3.676 & 0.467 & 3.135 & 0.323 & 3.000 & 3.605 \\
\bottomrule
\end{tabular}%
}
\end{table*}

\section{Experiments}
This section reports the experiment setup and results.
Further details(including the robustness and sensitivity of the method) are provided in Appendices~\ref{sec:appendix_details}.

\subsection{Experimental Setup}
Evaluation employs a multi-model strategy using LLMs, including GPT-4, GPT-oss-120b, Gemini-2.5-Pro, and DeepSeek-V3.2. Our diverse model selection reduces bias, ensuring findings reflect generalized capabilities. The specific evaluation metrics and baseline methods are detailed below.

\paragraph{Evaluation Metrics.}
To implement the PACE evaluation framework, we report five quantitative metrics.
For Population Alignment, we calculate \textbf{JSD} and \textbf{KL} to evaluate the distribution fidelity of the population dimension, while using diversity error (\textbf {DivErr}) to quantify structural alignment errors.
For Sociological Consistency, we score Archetypal Relevance (\textbf{ArchRel}) and Individual Consistency (\textbf{IndCon}) on a scale of 1-5 points. We use the LLM-as-a-judge method as the paradigm~\cite{llm_as_judges}, and the reliability of this automated judge was also verified through human evaluation.

\paragraph{Baselines.}
We compare \textbf{HAG(Our)} against four baselines categorized into two paradigms, specifically selected for topic-dependent population generation tasks.
Within Data-based Retrieval, \textbf{Random Select} serves as a topic-agnostic baseline by randomly sampling personas from the WVS pool, while \textbf{Topic-Retrieval} selects the Top-$N$ personas based on semantic similarity by embedding the topic and each WVS user's text using a sentence transformer model. 
Within LLM-based Generation, \textbf{LLM Generate} directly generates personas via end-to-end prompting with the topic and persona template. Additionally, we include \textbf{HAG-Flat}, an ablation variant that independently generates each demographic dimension distribution (conditioned topic only) to isolate the contribution of our hierarchical dependency modeling.

\begin{figure}[t]
  \centering
  \includegraphics[width=0.9\linewidth, height=5.7cm ]{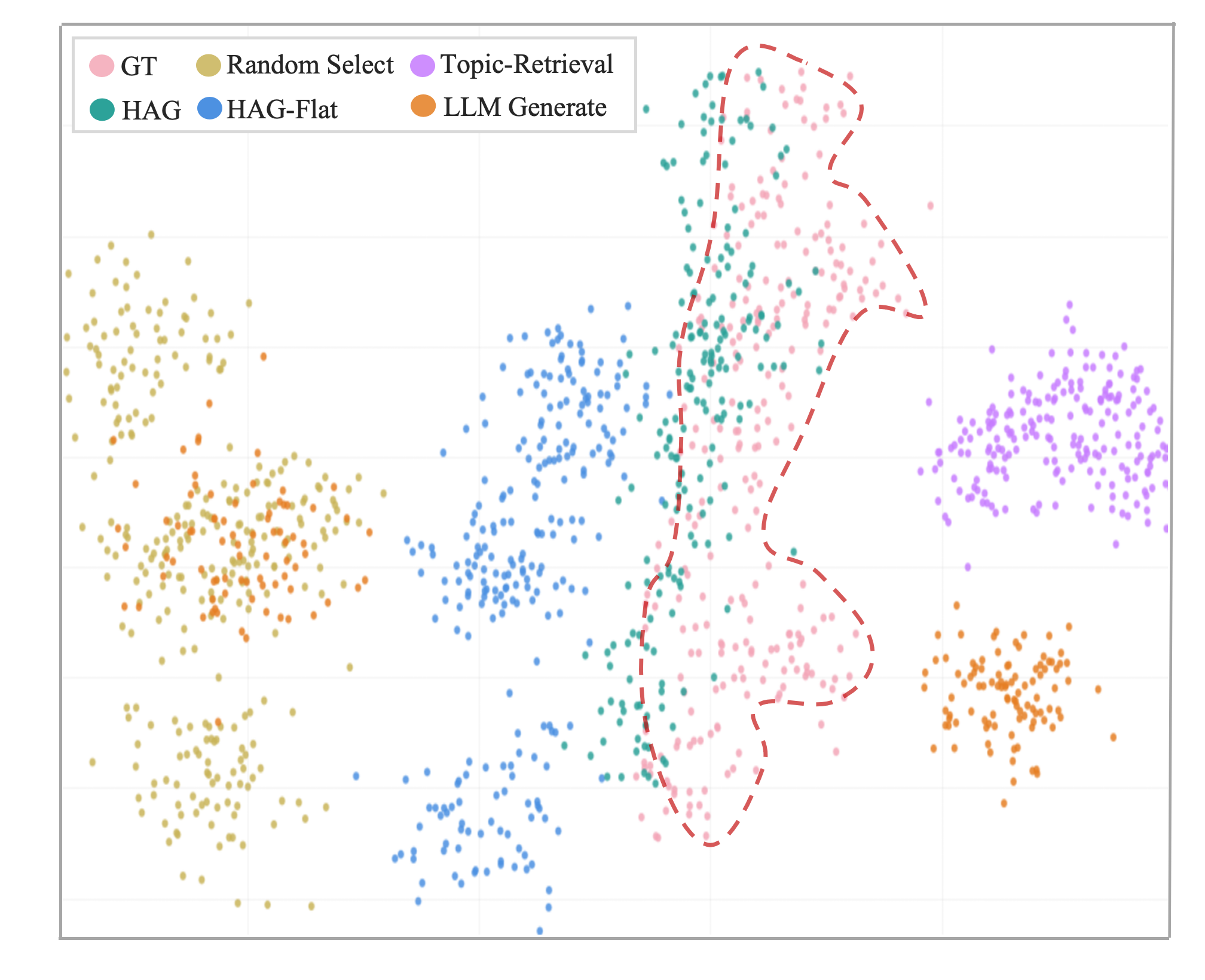}
  \caption{Visualization of the Population Manifold Structure in the Latent Persona Embedding Space (t-SNE).The area enclosed by the red dashed lines delineates the target mapping region of the Ground Truth population.}
  \label{fig.tsne}
\end{figure}

\subsection{Main Results}

Table~\ref{tab:main_results} shows the comparison across three domains. Overall, HAG achieves comprehensive improvements in both statistical alignment and semantic consistency. 
Specifically, compared to the LLM Generate baseline on the Bluesky dataset, HAG obtains an average improvement of \textbf{37.7\%} on population alignment metrics and \textbf{18.8\%} on sociological consistency metrics.
This superior performance trend is consistent across domains.

Detailed comparisons reveal that Data-based Retrieval methods consistently exhibit poor fidelity and consistency in two evaluation aspects, with Mean JSD scores exceeding 0.500 and IndCon often falling below 3.200 on most domains. Regarding structural alignment, both Random Select and LLM Generate suffer from high errors.
We also observed that Gemini-2.5-Pro exhibited significant anomalies on the Bluesky dataset, which inspection traced to world knowledge hallucinations regarding specific topics.
Notably, on the IMDB dataset, LLM Generate obtains a high DivErr (0.555), comparable to the noise of Random Select (0.535), but HAG reduces this to 0.322. These results confirm that HAG effectively preserves the complex structural heterogeneity of real-world populations. Furthermore, HAG consistently outperforms the HAG-Flat variant across all domains, validating the effectiveness of our hierarchical structure.

\begin{figure}[t]
  \centering
  \includegraphics[width=1\linewidth, height=5.5cm]{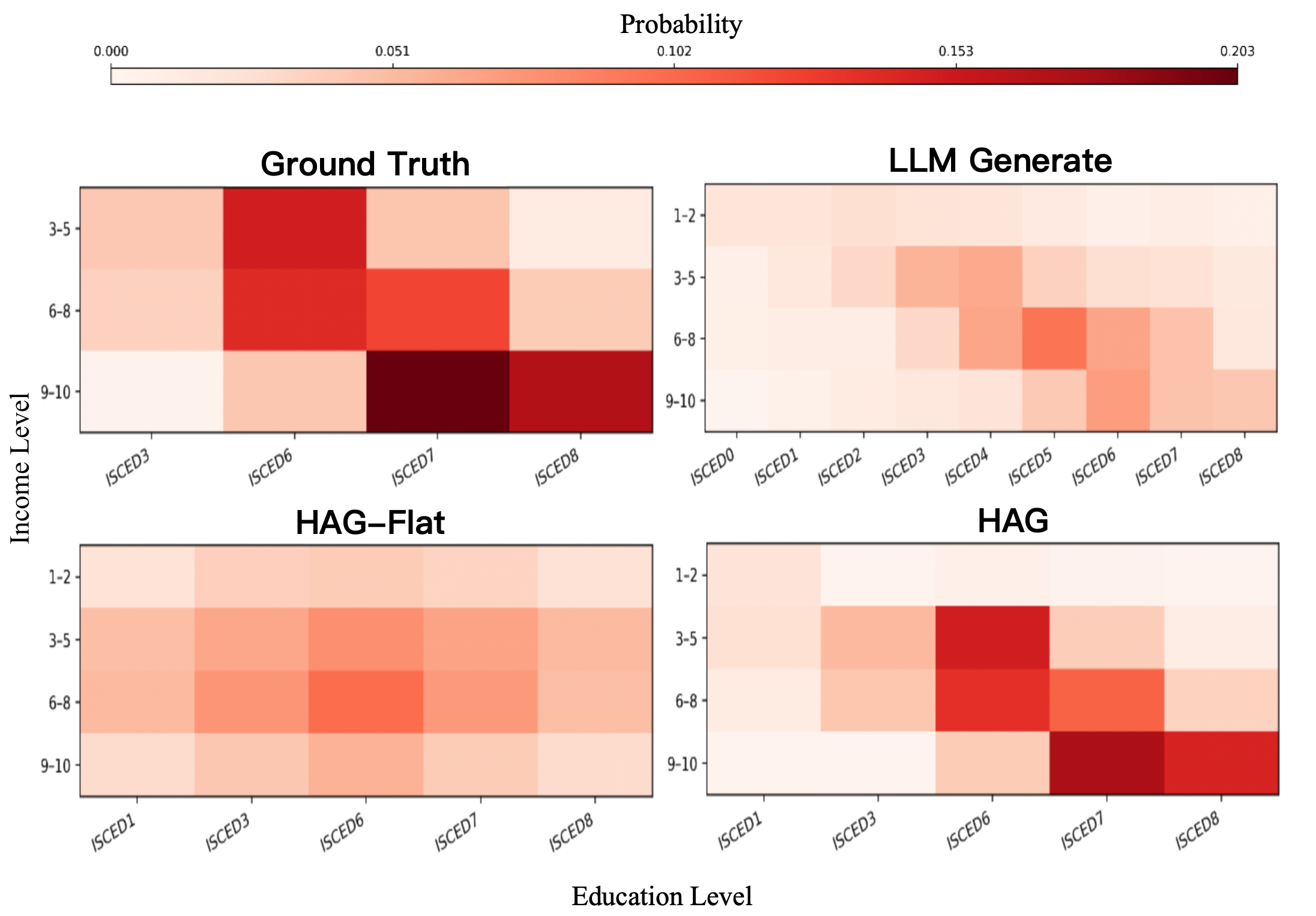}
  \caption{A joint distribution heatmap visualizing the correlation between education level and income level under the scientific research topic, including GT and three LLM-based Generation Methods.}
  \label{fig:heat}
\end{figure}

\subsection{Qualitative Analysis}
Quantitative metrics show HAG's superiority, and we further conduct qualitative analysis to investigate the intrinsic mechanisms. We answer two Research Questions (RQs).

\begin{figure}[t]
  \centering
  \includegraphics[width=1\linewidth]{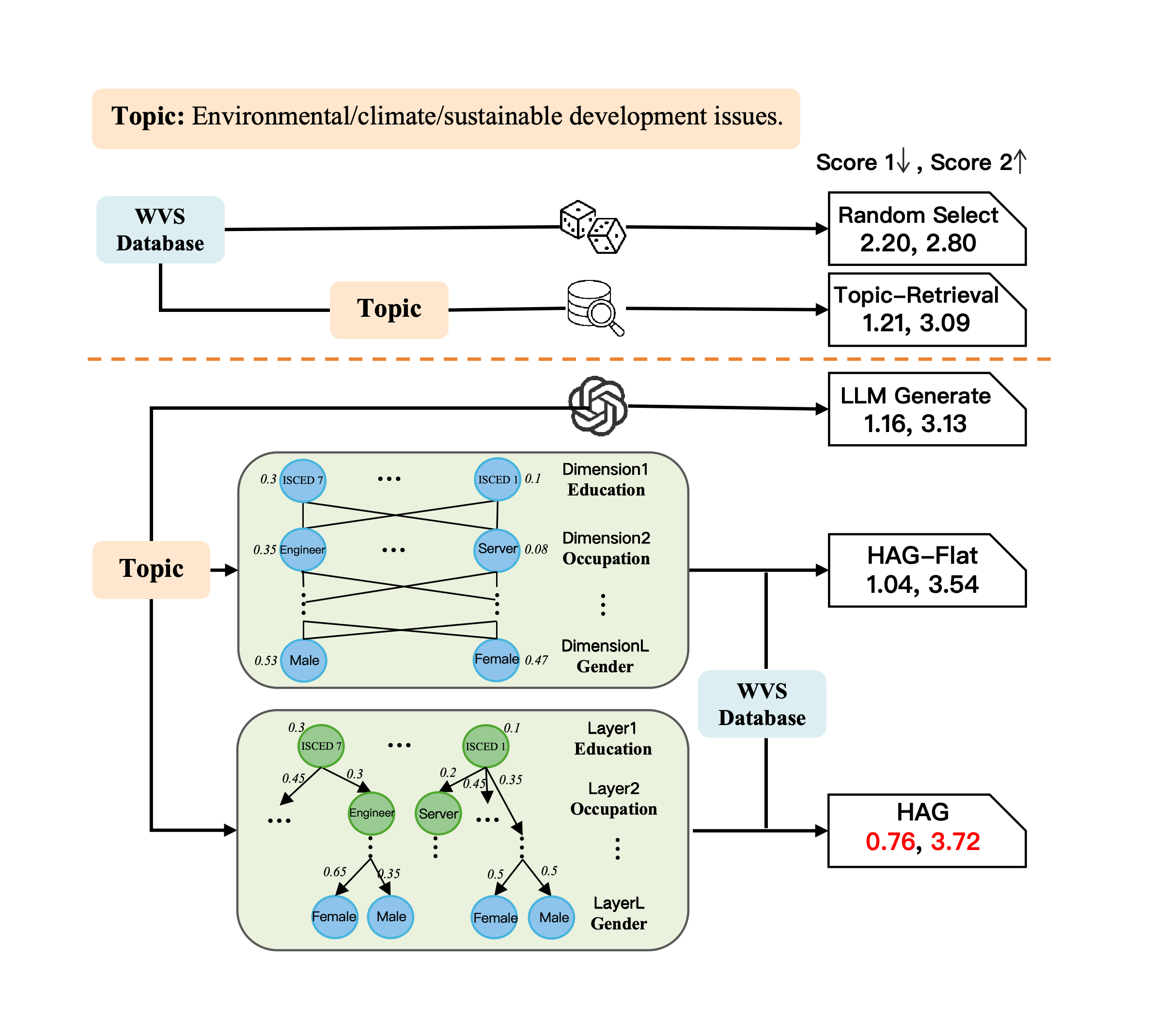}
  \caption{Demonstration of the case study.}
  \label{fig.case}
\end{figure}

\paragraph{RQ1: Does the population maintain a macro manifold structure consistent with reality?}
We upgrade our perspective from attribute statistics to the examination of the Population Manifold. We embed personas into high-dimensional dense vectors to capture their complex sociological features, and visualize their local clustering structure in a latent persona embedding space using t-SNE dimensionality reduction. 
Observed in Figure~\ref{fig.tsne}, data-based retrieval methods drift into distal regions, indicating a semantic misalignment with the target population. 
LLM Generate clusters tightly in a disjoint region, signaling severe mode collapse and a failure to capture the semantic diversity of the topic. 
Conversely, HAG demonstrates the highest degree of overlap with the GT, effectively filling the target manifold. This confirms that HAG can not only align marginal statistics but also successfully capture the complex, high-dimensional manifold structure of the real-world population.

\paragraph{RQ2: Does the population capture joint distributions and conditional dependencies?}
We visualize the correlation between education level and income level under scientific research topics through joint distribution heatmaps, shown in Figure~\ref{fig:heat}. 
LLM Generate produces a diffuse spread covering all education levels, including lower levels absent in the GT, indicating a failure to align with the demographic scope.
HAG-Flat exhibits a symmetric, grid-like distribution. Because it combines dimension values independently, leading to uniform probabilities.
Conversely, HAG closely matches the diagonal trend and high-density regions of the GT. This validates that our Topic-Adaptive Distribution Tree successfully locks in conditional dependencies, ensuring the generated populationconforms to the macro multi-attribute joint distribution, and that the attributes of micro individuals are also relevant and reasonable.

\subsection{Case Study}
Figure~\ref{fig.case} illustrates a case study on the topic of "Environmental". We trace the workflow of five methods and report their average performance on Alignment (Score 1) and Consistency (Score 2).
Retrieval methods get high alignment errors. LLM Generate acts as a black box with moderate performance. The HAG-Flat variant constructs personas by combining dimensions independently, which improves consistency but lacks structural precision.
HAG explicitly models conditional dependencies via a hierarchical demographic tree, whose structure enables HAG to achieve the best scores.

\subsection{Generalization and Adaptability Analysis}
\label{sec:generalization_analysis}
To evaluate adaptability on emerging topics (e.g., "Mars Colonization"), we analyze the trade-off between Structural Diversity (Gini-Simpson Index) and Sociological Consistency (defined as $S_{cons} = (\mathcal{\text{ArchRel}} + \mathcal{\text{IndCon}})/2$).

\begin{figure}[t]
  \centering
  \includegraphics[width=0.9\linewidth]{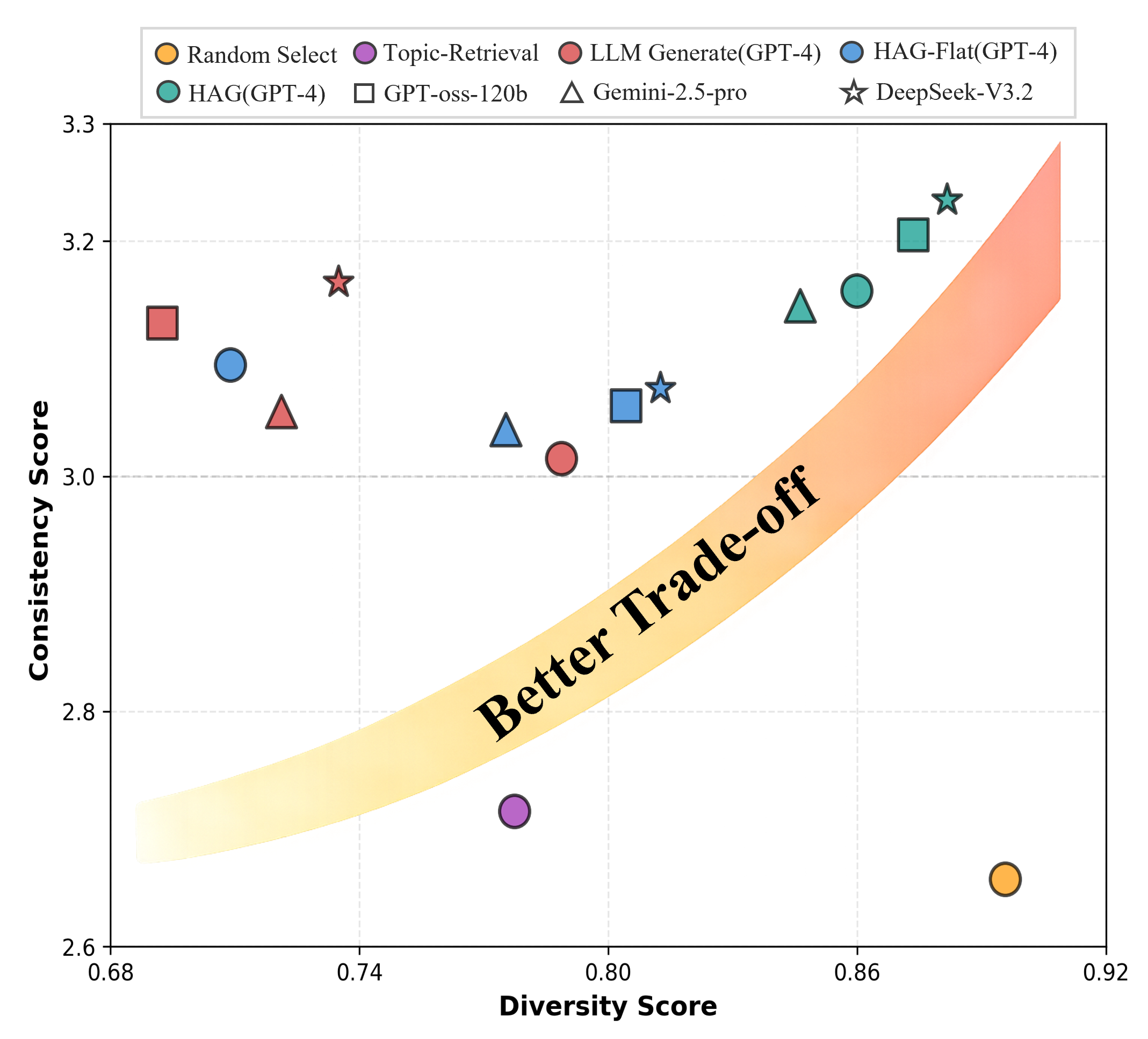}
  \caption{Diversity-Consistency Trade-off on Emerging Topic "Mars Colonization".}
  \label{fig.tradeoff}
\end{figure}

As illustrated in Figure~\ref{fig.tradeoff}, Random Select occupies the high-diversity but low-consistency region, representing mere statistical noise rather than meaningful heterogeneity. LLM Generate fluctuates in the middle, often succumbing to logical coherence to satisfy diversity. And the performance of LLM Generate and HAG-Flat is easily affected by the models.
In contrast, HAG is at the \textbf{Pareto Frontier} of this trade-off, maintaining high consistency without compromising population diversity. This confirms that by combining the Topic-Adaptive demographic distribution tree with grounded instantiation, HAG robustly generalizes to open-world scenarios, generating populations that are both diverse and sociologically rational.

\section{Related Work}
\paragraph{Persona generation.}
Persona generation turns a population distribution into concrete agent personas, and existing approaches largely follow either retrieval from real-user footprints or LLM-based generation. Retrieval-style pipelines typically build personas from real-world data: some retrieve or enrich existing user personas for a given topic~\cite{oasis, socioverse}, while others directly infer implicit personas from user logs~\cite{usp}. Generative methods generate personas in two ways: one infers personas from texts by extracting salient information under a certain topic~\cite{persona_hub, pop_aligned}; the other generates personas directly under predefined schemas such as persona templates or descriptions~\cite{yulan, promise_catch, mpg, llmgenerate}. However, personas produced by existing methods often fall short in two ways: they either lack a macro-level awareness of the topic or deviate from real-world population distributions, leading to biased persona generation.~\cite{promise_catch,pop_aligned}

\paragraph{Social simulation.}
LLM-driven social simulation replaces simple heuristic behavioral rules with LLM-based policies. In such cases, agent fidelity becomes a key determinant of simulation validity~\cite{fidelity}. Social media simulators either instantiate scenario-matched subpopulations from real-user pools~\cite{oasis,socioverse} or translate scenario descriptions into executable agents and environments~\cite{yulan} to reduce mismatches between intended and implemented behaviors, while recommendation simulators use users’ reactions to emulate how platforms decide what to show next~\cite{recagent,llmrec, feedbackloop}. Still, structural deviation (e.g., overestimated political homophily) can arise in social simulations, indicating the need for stronger fidelity and calibration checks~\cite{homophilybias,promise_catch}.

\section{Conclusion and Future Work}
We propose HAG, a Hierarchical Demographic Tree-based Agent Generation framework that combine Topic-Adaptive distribution tree with real-world grounded instantiation, which improves macro-level distribution fidelity and micro-level individual consistency in simulations. Furthermore, we establish a multi-domain benchmark and a comprehensive PACE evaluation framework. 
Future work will enrich the benchmark across diverse contexts and enhance the WKM with dynamic RAG.

\section*{Limitations}
Despite its effectiveness, HAG's performance is inherently bounded by the knowledge capacity of the underlying World Knowledge Model. The accuracy of the generated demographic tree depends on the World Knowledge Model's understanding of the specific topic. For instance, our experiments revealed that the Gemini-2.5-Pro model exhibited significant hallucinations on specific topics within the Bluesky dataset, leading to distributional anomalies. This observation underscores the critical importance of the World Knowledge Model's factual reliability. Additionally, as societal structures evolve, the static nature of the World Values Survey database may eventually introduce temporal lag, necessitating periodic updates or integration with real-time data streams to maintain fidelity.

\section*{Ethics Statement}
This research is conducted with strict adherence to ethical guidelines regarding data privacy and fairness.

\textbf{Data Privacy and Usage.} The real-world data of our framework relies exclusively on publicly available and anonymized datasets. Our method generates agents based on statistical joint distributions and aggregate patterns rather than reproducing specific identifiable individuals. HAG serves as a privacy-preserving alternative to using raw user logs for simulation, effectively mitigating the risk of re-identification or privacy leakage.

\textbf{Mitigation of Misuse.} We firmly condemn the misuse of this technology for generating disinformation, fake grassroots support, or malicious social manipulation. While HAG lowers the barrier for high-fidelity simulation, it also carries the risk of being deployed to simulate and exploit vulnerable populations. We urge researchers and practitioners to employ this framework solely for analytical and benevolent purposes.

\textbf{Bias and Fairness.} Researchers must remain aware that while HAG mitigates statistical biases through grounding, it cannot entirely eliminate residual biases inherited from the underlying LLMs or the survey data itself. There is a risk that historical biases present in the WVS data could be perpetuated in the synthesized population. Therefore, we recommend a "human-in-the-loop" approach for critical applications to audit generated agents for potential stereotyping or under-representation of marginalized groups before deployment.

\section*{Acknowledgment}
This work was supported by the Strategic Priority Research Program of the CAS (No. XDB0680302),
and the National Natural Science Foundation of China (No.U21B2046, No.62202448).

\bibliography{custom}
\newpage

\appendix

\section{Evaluation Details}
\label{sec:appendix_eval_details}

This section details the construction of our evaluation benchmark and the protocols for both automated and human assessment.

\subsection{Benchmark Construction}
\label{subsec:benchmark_construction}

To evaluate population generation in realistic scenarios, we constructed a Topic-Conditioned Demographic Benchmark derived from three datasets: Bluesky Social, Amazon Reviews 2023, and IMDb Movie Reviews.

\paragraph{Data Preprocessing and Topic Selection.}

For each dataset, we applied rigorous filtering to ensure the quality of the behavioral text used for demographic inference:
\begin{itemize}
    \item \textbf{Data Cleaning:} We filtered out short texts (fewer than 15 tokens) and removed non-natural language content (e.g., spam, URLs). To ensure an appropriate amount of context for profiling, we subsampled the review datasets (Amazon/IMDb) by filtering reviews within a specified time range and retaining users with an appropriate number of posted reviews. For Amazon Reviews 2023, we kept reviews posted in 2023 and retained users with at least 10 reviews. For IMDb Movie Reviews, we kept reviews posted in 2021--2022.
    \item \textbf{Topic Selection:} We selected topics based on two criteria: (1) \textit{Volume}, ensuring at least 50 unique users participated in the discussion; and (2) \textit{Controversy/Diversity}, ensuring the topic elicited a wide range of sentiments and perspectives. Guided by these criteria, we selected eleven discussion themes in Bluesky, two product categories in Amazon, and two movies in IMDb as our topics. The selected topics are detailed in Table~\ref{tab:topics}.
\end{itemize}

\begin{table*}[t]
    \centering
    \small
    \caption{Topics selected from each dataset.}
    \label{tab:topics}
    \renewcommand{\arraystretch}{1.1}
    \setlength{\tabcolsep}{4pt} 
    \begin{tabular}{p{0.22\textwidth} p{0.2\textwidth} p{0.56\textwidth}}
    \toprule
    \textbf{Dataset} & \textbf{Theme/Category/Movie} & \textbf{Topic} \\
    \midrule
    \multirow{11}{*}{Bluesky Social Dataset} 
        & \#Disability & People with disabilities and disability issues. \\
        & \#UkrainianView & Ukrainian perspectives and experiences during the war. \\
        & AcademicSky & Academic discussions, including higher education, academic work, academic discourse, etc. (mainly aimed at academic groups). \\
        & BlackSky & The voices and issues of black users (the group consists of individuals who identify themselves as black users). \\
        & BookSky & Reading, book recommendations, literature. \\
        & Game Dev & Game development topics, including production, programming, design, etc.. \\
        & GreenSky & Environmental/climate/sustainable development issues (such as climate change, emissions, energy, etc.). \\
        & News & Headline content released by news organizations (which may include major current events in various countries). \\
        & Political Science & Research and Discussion in the Field of Political Science/International Relations. \\
        & Science & Science communication, academic/research personnel, scientific topics and popular science content. \\
        & What’s History & Historians/Historical Topics: Historical Research, Historical Stories, Historical Figures. \\
    \midrule
    \multirow{2}{*}{Amazon Reviews 2023} 
        & Baby products & Baby products such as diapers, feeding supplies, and baby care essentials. \\
        & Musical Instruments & Musical instruments and related gear such as guitars, keyboards, drums, and recording equipment. \\
    \midrule
    \multirow{2}{*}{IMDB Movies User Reviews} 
        & Forrest Gump & Forrest Gump (1994) \\
        & Joker & Joker (2019) \\
    \bottomrule
    \end{tabular}
\end{table*}

\paragraph{Text-to-Persona Pipeline.}
Since public datasets lack fine-grained demographic labels, we employed a Text-to-Persona approach to infer Ground Truth personas from user-generated content. The attributes of personas are derived from the demographic attributes extracted from the WVS dataset, as shown in Table~\ref{tab:wvs_attributes}. 
We utilized \textbf{GPT-4o} as the inference engine due to its superior reasoning capabilities. The model accepts a user's historical posts/reviews as input and infers demographic attributes (Age, Gender, Education, Income, Occupation, Social Class). To minimize hallucinations, the model was instructed to return "Unknown" if the text provided insufficient clues.

\begin{table}[t]
    \centering
    \small 
    \caption{The schema of demographic attributes extracted from the WVS dataset.}
    \label{tab:wvs_attributes}
    \setlength{\tabcolsep}{4pt}
    \begin{tabular}{l l l} 
    \toprule
    \textbf{Category} & \textbf{Dimension} & \textbf{WVS Code} \\
    \midrule
    \multirow{5}{*}{Basic Demographics} 
        & Country & \texttt{B\_COUNTRY} \\
        & Language & \texttt{S\_INTLANGUAGE} \\
        & Gender & \texttt{Q260} \\
        & Age & \texttt{Q262} \\
        & Marital Status & \texttt{Q273} \\
    \midrule
    \multirow{5}{*}{Socio-Economic Status} 
        & Education & \texttt{Q275} \\
        & Occupation & \texttt{Q281} \\
        & Income Level & \texttt{Q288} \\
        & Financial Status & \texttt{Q286} \\
        & Social Class & \texttt{Q287} \\
    \midrule
    \multirow{2}{*}{Cultural Identity} 
        & Religion & \texttt{Q289} \\
        & Ethnicity & \texttt{Q290} \\
    \bottomrule
    \end{tabular}
\end{table}

\subsection{Automated Evaluation Setup}
For the \textit{Sociological Consistency} metric within our PACE framework, we adopted the \textbf{LLM-as-a-Judge} paradigm.
We employed \textbf{GPT-4o} as the evaluator to score generated agents on two dimensions: Archetypal Relevance (ArchRel) and Individual Consistency (IndCon). The judge evaluates agents on a 5-point Likert scale, providing reasoning for each score. 

\subsection{Human Verification Protocol}
\label{subsec:human_verification}

To ensure the validity of our evaluation pipeline, we conducted a rigorous human verification process. This process served two purposes:
(1) Benchmark Validation: Verifying the accuracy of the GPT-4o inferenced personas (Ground Truth). (2) Judge Reliability: Verifying the alignment between the automated LLM judge's scores and human sociological assessment.

\paragraph{Expert Annotators.}
We recruited 10 PhD candidates in Sociology from universities to serve as expert annotators. All annotators possessed advanced knowledge of quantitative research methods and social stratification. They were given the same detailed instructions as the prompts we used when evaluating via LLM-as-a-judge to ensure consistency. Participants were financially compensated for their time, with a payment rate determined in accordance with the standard stipend for graduate research assistants.

\paragraph{Adaptive Sampling Strategy.}
\label{para:sampling_strategy}

In both benchmark construction and experimental generation, the total population size $M$ varies significantly across different topics (e.g., niche discussions may contain fewer than 100 users, while broad social topics involve thousands). A fixed sample size would thus be statistically invalid: it would either undersample large populations or oversample small ones. To address this, we employed an Adaptive Random Sampling strategy based on finite population correction.

The sample size $n$ is calculated as follows:
\begin{equation}
n = \min\left(M, \max\left(30, \frac{n_0}{1 + \frac{n_0 - 1}{M}}\right)\right)
\end{equation}
\begin{equation}
n_0 = \left(\frac{Z \cdot \sigma}{E}\right)^2
\end{equation}
where $M$ is the total population size for a specific topic, $n_0$ is the initial sample size for an infinite population, $Z$ is the Z-score corresponding to the confidence level (1.96 for 95\% confidence), $\sigma$ is the estimated standard deviation (set to 1.0 as a conservative estimate for Likert-scale variance), and $E$ is the acceptable margin of error (set to $\pm 0.2$). This formula first calculates the initial sample size based on the infinite population assumption, then adjusts it using the finite population correction factor, ensuring that the final sample size is neither less than 30 (to maintain statistical reliability) nor exceeds the total population size $M$.

The specific sampling statistics derived from this protocol are detailed in Table~\ref{tab:sampling_protocol}.

\begin{table}[h]
    \centering
    \small
    \caption{Adaptive Sampling Protocol Statistics.}
    \label{tab:sampling_protocol}
    \renewcommand{\arraystretch}{1.1}
    \setlength{\tabcolsep}{4pt} 
    \begin{tabular}{@{}rccc@{}}
    \toprule
    \textbf{Pop. ($M$)} & \textbf{Sample ($n$)} & \textbf{Prop. (\%)} & \textbf{Error ($E$)} \\
    \midrule
    20   & 20 & 100.00 & 0.00 \\
    30   & 30 & 100.00 & 0.00 \\
    31   & 30 & 96.77  & 0.07 \\
    100  & 50 & 50.00  & 0.20 \\
    101  & 50 & 49.50  & 0.20 \\
    500  & 81 & 16.20  & 0.20 \\
    501  & 81 & 16.17  & 0.20 \\
    1000 & 88 & 8.80   & 0.20 \\
    1001 & 88 & 8.79   & 0.20 \\
    2000 & 92 & 4.60   & 0.20 \\
    \bottomrule
    \end{tabular}
\end{table}

Results from this verification showed a\textbf{ 92\%} agreement rate between the expert annotators and the GPT-4o derived benchmark/scores, validating the reliability of our automated pipeline.

\paragraph{Internal validation protocols.}
Beyond our human-concordance check, we implemented internal validation protocols:
\begin{itemize}
    \item \textbf{Signal-Dependency Test (Masking):} To test if the LLM hallucinates unobserved traits, we sampled user texts containing explicit demographic declarations (e.g., stated Income Level or Religion). We then ran the extraction on a masked version of the text where these specific signals were artificially removed. The pipeline correctly abstained (outputting "Unknown" rather than hallucinating a guess) in 94.3\% of the masked cases. This proves the extraction is strictly contingent on the presence of textual evidence, not on latent LLM priors.
    \item \textbf{Repeatability Test (Stability):} To ensure the model wasn't relying on unstable, ad-hoc heuristics, we ran the inference pipeline three independent times on the same set of user texts. The inferred personas demonstrated a 95.8\% consistency rate across the three trials.
\end{itemize}

\section{Experimental Details}
\label{sec:appendix_details}

This appendix provides additional details regarding the experimental setup, including embedding model selection and parameter settings for our proposed HAG framework.

\subsection{Embedding Model}
\label{sec:embedding_model}

For retrieval and clustering tasks in baselines and evaluation stage, we used the \texttt{sentence-transformers/all-MiniLM-L6-v2}~\footnote{\url{https://huggingface.co/sentence-transformers/all-MiniLM-L6-v2}} model to encode the input topic and all personas from the processed WVS dataset and the generated population into dense embeddings. We set the batch size to 1 when encoding the topic and to 32 otherwise. These embeddings were used to compute cosine similarity between the topic and user personas in the Topic-Retrieval baseline and to perform K-means clustering for the Archetypal Relevance metric.

\subsection{Method Parameters}
\label{sec:HAG_params}

Our proposed HAG framework sets specific parameters for constructing the Topic-Adaptive demographic distribution tree. These parameters were fixed across our experiments as follows:
\begin{itemize}
    \item Maximum number of prioritized dimensions  ($max\_depth$): 5
    \item Maximum number of values per dimension ($max\_branches$): 5
    \item Sampling temperature of the world knowledge model ($temperature$): 0.0
\end{itemize}

\subsection{Robustness to Data Sparsity: HIT vs. MISS Analysis}
\label{appendix:hit_miss_analysis}

To evaluate the robustness of our framework against data sparsity, particularly in highly niche domains, we analyze the HIT (successful matching with empirical World Values Survey users) versus MISS (reliance on agentic augmentation) ratios during the instantiation phase. Table~\ref{tab:hit_miss_ratios} details these statistics across four distinct themes from our benchmark: BlackSky, Game Dev, AcademicSky, and GreenSky.

The statistical distribution reveals that even for highly specialized or niche topics, the HIT rates consistently dominate the generation process (ranging from 64.74\% to 68.65\%). Notably, there is no significant numerical degradation in the HIT ratio for these niche communities compared to more general or popular topics. This empirical evidence demonstrates that our hierarchical conditional system successfully handles data sparsity without over-relying on the LLM-driven augmentation step, thereby maintaining strong empirical sociological grounding across diverse simulation scenarios.

\begin{table}[h]
    \centering
    \setlength{\tabcolsep}{8pt}
    \caption{HIT and MISS ratios across different topics.}
    \label{tab:hit_miss_ratios}
    \begin{tabular}{lcc}
        \hline
        Topic & HIT Ratio & MISS Ratio \\
        \hline
        BlackSky    & 68.37\% & 31.63\% \\
        Game Dev    & 68.65\% & 31.35\% \\
        AcademicSky & 68.47\% & 31.53\% \\
        GreenSky    & 64.74\% & 35.26\% \\
        \hline
    \end{tabular}
\end{table}

\subsection{Sensitivity to the Distribution Tree Structure}

\noindent\textbf{Sensitivity to Tree Depth:} We evaluated varying tree depths (3, 6, and 9) and observed a clear trade-off curve across both macro and micro metrics, as summarized in Table~\ref{tab:tree_depth}. A shallow tree (depth=3) lacks the expressiveness to capture complex joint distributions, resulting in the poorest macro-distribution Alignment (0.875, where lower is better) and the lowest micro-level Consistency (3.56), despite having a high initial HIT Ratio (79.49\%). The appropriate setting (depth=6) strikes the best balance, achieving the strongest Alignment (0.5041) and the highest Consistency (3.71) while maintaining a healthy WVS HIT Ratio (66.74\%). Conversely, an overly deep tree (depth=9) imposes excessively strict constraints. As the conditional path grows longer, the database struggles to find exact matches (data sparsity), causing the HIT Ratio to plummet to 24.55\%. This forces the system to rely heavily on ``Agentic Augmentation,'' triggering error propagation that degrades both Alignment (0.5822) and Consistency (3.62).

\begin{table}[h]
    \centering
    \small
    \setlength{\tabcolsep}{4pt}
    \caption{Performance metrics across varying tree depths.}
    \label{tab:tree_depth}
    \begin{tabular}{lccc}
        \hline
        Depth & HIT Ratio & Alignment ($\downarrow$) & Consistency ($\uparrow$) \\
        \hline
        3 & 79.49\% & 1.575 & 3.56 \\
        6 & 66.74\% & \textbf{0.5041} & \textbf{3.71} \\
        9 & 24.55\% & 0.5822 & 3.62 \\
        \hline
    \end{tabular}
\end{table}

\noindent\textbf{Sensitivity to Dimension Selection:} To evaluate the selection of dimensions, we compared our HAG method (Topic-Adaptive Tree) against a baseline Fixed-Dimension Tree. In the HAG method, the WKM dynamically selects dimensions and priorities. For example, for the AcademicSky theme in the Bluesky benchmark, the WKM dynamically selected \textbf{Education $\rightarrow$ Occupation $\rightarrow$ Country $\rightarrow$ Language $\rightarrow$ Age}. Conversely, the Fixed-Dimension Tree enforces the use of a fixed and universal set of sociological dimensions: \textbf{Age $\rightarrow$ Gender $\rightarrow$ Country $\rightarrow$ Education $\rightarrow$ Income}. As shown in Table~\ref{tab:dim_selection}, the adaptive HAG tree significantly outperforms the fixed tree in terms of alignment (0.573 vs.\ 1.205) and consistency (3.63 vs.\ 3.50). This proves that fixed trees waste hierarchical capacity on topic-irrelevant attributes while omitting key features, confirming that dynamic dimension selection is crucial for maintaining topic adaptability.

\begin{table}[h]
    \centering
    \small
    \setlength{\tabcolsep}{4pt}
    \caption{Comparison of dynamic vs. fixed dimension selection.}
    \label{tab:dim_selection}
    \begin{tabular}{lcc}
        \hline
        Method & Alignment ($\downarrow$) & Consistency ($\uparrow$) \\
        \hline
        Fixed-Dimension & 1.205 & 3.50 \\
        HAG (Topic-Adaptive) & \textbf{0.573} & \textbf{3.63} \\
        \hline
    \end{tabular}
\end{table}

In summary, these empirical results confirm our intuition: a carefully balanced tree depth is critical to maximize joint distribution expressiveness without triggering data sparsity and error propagation, while dynamic dimension selection is vital to capture topic-specific sociological features. Together, these findings strongly validate the necessity of our WKM-driven, topic-adaptive tree construction approach.

\clearpage
\onecolumn
\section{Prompts}
\label{app:prompt}


\begin{UnifiedPromptBox}{Prompt for identifying prioritized topic-relevant dimensions}
{\ttfamily
You are a computational sociologist. Your task is to determine the most important user profile dimensions for a social network simulation on the topic "\{topic\}".

\vspace{2mm}
Please identify up to 5 most critical demographic dimensions from the table below and rank them in descending order of their influence on people's opinions and behaviors related to this topic.

\vspace{2mm}
The dimensions in the table are:\\
1. Gender\\
2. Age\\
3. Education\\
4. Country\\
5. Language\\
6. Marital status\\
7. Occupation\\
8. Financial status\\
9. Social class\\
10. Income level\\
11. Religion\\
12. Ethnicity

\vspace{2mm}
Please output a list in the following JSON format strictly: \\
\{\{ \\
  "dimensions": ["Dimension 1", "Dimension 2", "Dimension 3", "Dimension 4"] \\
\}\} \\
}
\end{UnifiedPromptBox}


\begin{UnifiedPromptBox}{Prompt for generating a conditioned distribution over a demographic dimension}
{\ttfamily
You are a computational sociologist. \{context\_str\}, please generate a plausible probability distribution for the dimension "\{dimension\}".

\vspace{2mm}
List the primary values for this dimension and assign a probability to each.

\vspace{2mm}
Provide the most relevant and meaningful values for this context - you can provide anywhere from 1 to \{max\_branches\} values, depending on what makes sense for the given context.

\vspace{2mm}
\textbf{IMPORTANT}: Choose the number of values based on what is actually meaningful and significant for this specific context.\\
- If only 1-3 categories are truly relevant, use only 1-3 values.\\
- If more categories are meaningful, you can use up to \{max\_branches\} values.\\
- Do NOT artificially inflate the number of categories just to reach the maximum.

\vspace{2mm}
Focus on the most significant categories rather than trying to fill up to the maximum number.\\
The sum of all probabilities must be exactly 1.0.

\vspace{2mm}
Strictly adhere to the following JSON format for your output: \\
\{\{ \\
  "distribution": [ \\
    \{\{"value": "Value 1", "probability": 0.xx\}\}, \\
    ... \\
  ] \\
\}\} \\

\vspace{2mm}
\{allowed\_clause\}\\
}
\end{UnifiedPromptBox}

\begin{UnifiedPromptBox}{Prompt for Text-to-Persona Pipeline}
{\ttfamily
You are a computational sociologist analyzing social media posts to generate realistic user profiles.\\

\textbf{TASK}: Generate a user profile based on the provided social media posts from the "\{theme\}" community. \{dimension\_info\}\\
\textbf{USER'S POSTS}:\\
\{user\_text\}\\

\textbf{INSTRUCTIONS}:\\
1. Analyze the user's posts to infer their demographic characteristics. Make reasonable inferences based on the content, language, and context of the posts.\\
2. Generate a realistic user profile that matches the template structure. Only generate values for the dimensions specified in the template.\\
3. Replace all "\_\_FILL\_\_" placeholders with appropriate values. Ensure all values are chosen from the allowed constraints below.\\

\textbf{ALLOWED VALUES (choose ONLY from these lists)}:\\
\{constraints\_text\}\\

\textbf{IMPORTANT CONSTRAINTS}:\\
- You MUST choose values ONLY from the allowed lists above. Do NOT invent new values or categories.\\
- If you cannot determine a value from the posts, choose the most common/general option from the allowed list.\\
- Consider the theme context: "\{theme\}" community members may have specific characteristics.\\
- Only generate the dimensions specified in the template - do not add extra fields.\\

\textbf{OUTPUT FORMAT}:\\
Return a JSON object that exactly matches this template structure:\\
\{template\_json\}\\

\textbf{ANALYSIS GUIDELINES}:\\
- Age: Infer from language style, references to life events, generational markers.\\
- Education: Consider vocabulary, topic complexity, academic references.\\
- Country: Look for location mentions, cultural references, language patterns.\\
- Occupation: Analyze professional topics, work-related discussions.\\
- Religion: Consider religious references, holidays, cultural practices.\\
- Other fields: Make reasonable inferences based on available information.\\

Generate the user profile now:\\
}
\end{UnifiedPromptBox}


\begin{UnifiedPromptBox}{Evaluation prompt for Archetype Relevance}
{\ttfamily
You are an expert computational sociologist.\\

DOMINANT CLUSTERS:\\
\{dom\_snippet\}\\

THEME / TOPIC CONTEXT:\\
\{theme\_context\}\\

Question:\\
Are these dominant archetypes (typical groups) the core stakeholders for this topic?\\
Consider whether age, education, occupation, country, language and other demographics\\
form plausible and meaningful typical user types that align with sociological expectations.\\

Scoring guide:\\
- 1: Archetypes are completely irrelevant or implausible for this topic\\
- 3: Archetypes are somewhat relevant but have some issues\\
- 5: Archetypes are highly relevant and plausible as core stakeholders for this topic\\

Return format (must be a valid JSON object):\\
\{\{\\
"archetype\_coherence\_score": <int 1--5>,\\
"reasoning": "<short explanation>"\\
\}\}\\
}
\end{UnifiedPromptBox}


\begin{UnifiedPromptBox}{Evaluation prompt for Individual Consistency}
{\ttfamily
You are a computational social scientist who studies population structure.\\

Given the following theme/topic and a single agent profile, evaluate whether the\\
combination of demographic attributes (such as age, education, occupation, country, etc.)\\
is internally consistent and realistic for that theme.\\

Topic:\\
\{context\}\\

Agent Profile (JSON):\\
\{user\_profile\}\\

Your task:\\
- Please only judge from the perspective of ``logical consistency'' and give a rating of 1--5 to indicate whether the attribute combination of the agent is reasonable and consistent in this real-world topic:\\
\ \ \ \ 1 = Very unreasonable (with obvious contradictions)\\
\ \ \ \ 3 = Generally reasonable (with some doubts but acceptable)\\
\ \ \ \ 5 = Very reasonable (there is no obvious contradiction)\\

Return format (must be a valid JSON object):\\
\{\{\\
\ \ ``internal\_consistency\_score'': <int 1--5>,\\
\ \ ``reasoning'': ``<Short text explaining the main reason for judgment>''\\
\}\}\\
}
\end{UnifiedPromptBox}
\clearpage
\twocolumn

\end{document}